\documentclass[runningheads]{llncs}

\usepackage{eccv}

\usepackage{eccvabbrv}

\usepackage{graphicx}
\usepackage{booktabs}
\usepackage{algorithm}
\usepackage{multirow}
\usepackage{multicol}
\usepackage{booktabs}  
\usepackage{threeparttable}
\usepackage{amsfonts}       % blackboard math symbols
\usepackage{amssymb}     
\usepackage{amsmath}   
\usepackage{svg}
\usepackage{nicefrac}       % compact symbols for 1/2, etc.
\usepackage{microtype}      % microtypography
\usepackage{xcolor}         % colors
\usepackage{bm}     
\usepackage{xfrac} 
\usepackage{array}
\usepackage{makecell}
\usepackage{wrapfig}
\usepackage{enumitem}
\usepackage{algorithm}
\usepackage{algpseudocode}
\usepackage[utf8]{inputenc}
\newcommand{\PreserveBackslash}[1]{\let\temp=\\#1\let\\=\temp}
\newcolumntype{C}[1]{>{\PreserveBackslash\centering}p{#1}}
\newcolumntype{R}[1]{>{\PreserveBackslash\raggedleft}p{#1}}
\newcolumntype{L}[1]{>{\PreserveBackslash\raggedright}p{#1}}
\newcommand{\bihan}[1]{\textcolor{black}{#1}}

\usepackage[accsupp]{axessibility}  % Improves PDF readability for those with disabilities.

\usepackage[pagebackref,breaklinks,colorlinks,citecolor=eccvblue]{hyperref}
\usepackage{orcidlink}
\usepackage{wrapfig}
\usepackage{lipsum}

\begin{document}

\title{Make a Cheap Scaling: A Self-Cascade Diffusion \\ Model for Higher-Resolution Adaptation} 

\titlerunning{Make a Cheap Scaling}

\author{Lanqing Guo\inst{1,2}$^{\dag}$\orcidlink{0000-0002-9452-4723} \and
    Yingqing He\inst{2,3}$^\dag$\orcidlink{0000-0003-0134-8220}\and
    Haoxin Chen\inst{2}\orcidlink{0009-0000-6085-2107}\and
	Menghan Xia\inst{2}\orcidlink{0000-0001-9664-4967} \and
	Xiaodong Cun\inst{2}\orcidlink{0000-0003-3607-2236} \and
	  Yufei Wang\inst{1}\orcidlink{0000-0002-6326-7357} \and
        Siyu Huang\inst{4}\orcidlink{0000-0002-2929-0115} \and
     Yong Zhang\inst{2}$^{*}$\orcidlink{0000-0003-0066-3448} \and
     Xintao Wang\inst{2}\orcidlink{0000-0001-6585-8604} \and
          Qifeng Chen\inst{3}\orcidlink{0000-0003-2199-3948} \and
     Ying Shan\inst{2}\orcidlink{0000-0001-7673-8325} \and
  Bihan Wen\inst{1}$^{*}$\orcidlink{0000-0002-6874-6453
}}
\authorrunning{L. Guo et al.}

  \institute{
Nanyang Technological University 
\and
Tencent AI Lab \and The Hong Kong University of Science and Technology \and
 Clemson University \\
{\tt\small Project page: \url{https://guolanqing.github.io/Self-Cascade/}}}
% \author{Lanqing Guo\inst{1}\orcidlink{0000-1111-2222-3333} \and
% Second Author\inst{2,3}\orcidlink{1111-2222-3333-4444} \and
% Third Author\inst{3}\orcidlink{2222--3333-4444-5555}}

% % TODO FINAL: Replace with an abbreviated list of authors.

% % TODO FINAL: Replace with your institution list.
% Springer Heidelberg, Tiergartenstr.~17, 69121 Heidelberg, Germany

\maketitle

\let\thefootnote\relax\footnotetext{$^\dag$ Equal Contributions}
\let\thefootnote\relax\footnotetext{$^*$ Corresponding Authors}
\begin{abstract}
  Diffusion models have proven to be highly effective in image and video generation; 
  %however, they still face composition challenges when generating images of varying sizes due to single-scale training data. 
  \bihan{however, they encounter challenges in the correct composition of objects when generating images of varying sizes due to single-scale training data.}
  Adapting large pre-trained diffusion models \bihan{to} higher resolution demands substantial computational and optimization resources, yet achieving generation \bihan{capabilities} comparable to low-resolution models remains \bihan{challenging}.
  This paper proposes a novel self-cascade diffusion model
that leverages the knowledge gained from a well-trained low-resolution image/video generation model, 
enabling rapid adaptation to higher-resolution generation.
\bihan{Building on this}, we employ the pivot replacement strategy to \bihan{facilitate} a tuning-free version by progressively leveraging reliable semantic guidance derived from the low-resolution model.
We further propose to integrate a sequence of learnable multi-scale upsampler modules for a tuning version \bihan{capable of} efficiently \bihan{learning} structural details at a new scale from a small amount of newly acquired high-resolution training data.
% We further propose a pivot-guided noise re-schedule strategy to speed up the inference process and improve local structural details.
Compared to full fine-tuning, our approach achieves a $5\times$ training speed-up and requires only 0.002M tuning parameters. Extensive experiments demonstrate that our approach can quickly adapt to higher-resolution image and video synthesis by fine-tuning for just $10k$ steps, with virtually no additional inference time.
  \keywords{Image Synthesis \and Video Synthesis \and Diffusion Model \and Higher-Resolution Adaptation}
\end{abstract}

\section{Introduction}
\label{sec:intro}

\begin{figure}[!t]
\centering
% \vspace{-5mm}
% \includegraphics[width=.32\linewidth]{img/21_4.png}\hfill%
% \includegraphics[width=.32\linewidth]{img/90_4.png}\hfill%
% \includegraphics[width=.32\linewidth]{img/99_2.png} \\
% \includegraphics[width=.32\linewidth]{img/23_4_high.png}\hfill%
% \includegraphics[width=.32\linewidth]{img/91_4_high.png}\hfill%
% \includegraphics[width=.32\linewidth]{img/100_2_high.png}
\vspace{-0.3cm}
\includegraphics[width=.6\linewidth]{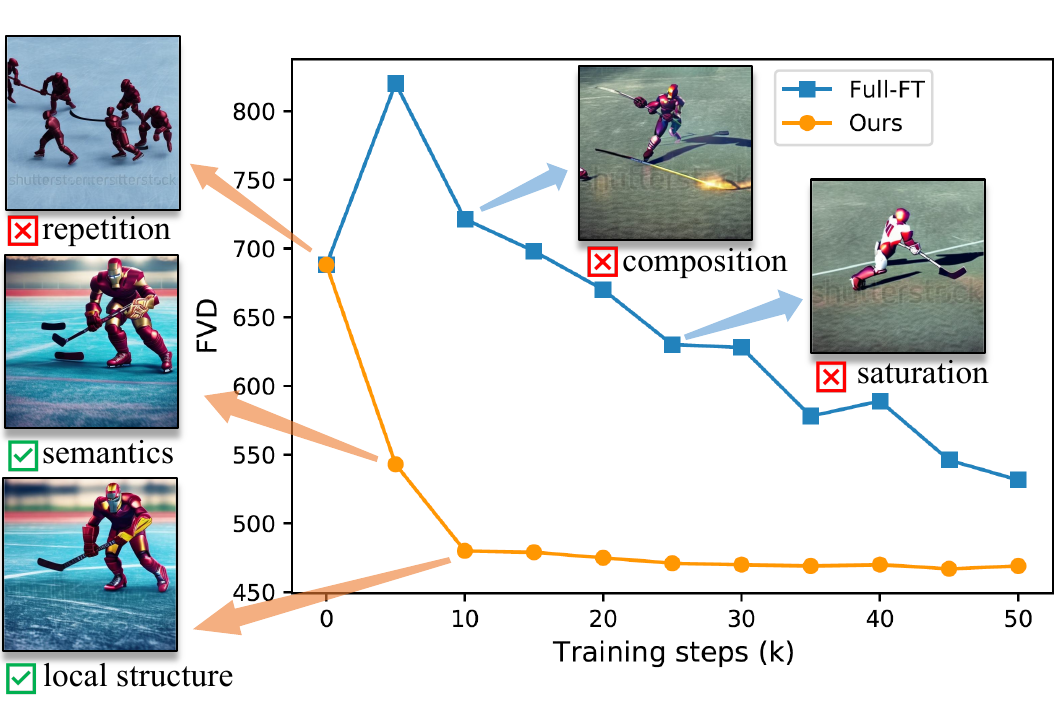} 
\vspace{-3mm}
\caption{
% The average FVD$\downarrow$ score on the \textit{Webvid-10M}~\cite{webvid} benchmark for both the full fine-tuning (Full-FT) and our proposed fast adaptation method (Ours) is assessed every $5k$ iterations. 
The FVD$\downarrow$ score averages for both the full fine-tuning (Full-FT) and our proposed fast adaptation method (Ours) are assessed every $5k$ iterations on the \textit{Webvid-10M}~\cite{webvid} benchmark.
We observe that full fine-tuning necessitates a large number of training steps and suffers from poor composition ability and desaturation issues. In contrast, our method enables rapid adaptation to the higher-resolution domain while preserving reliable semantic and local structure generation capabilities.}
\vspace{-0.2cm}
\label{fig:intro} 
\end{figure}

%In the past two years, stable diffusion (SD)~\cite{sd2-1-base,sdxl} has been a surge of interest in generative models, attracting both academic and industry attention.
\bihan{Over the past two years, stable diffusion (SD)~\cite{sd2-1-base,sdxl} has sparked great interest in generative models, gathering attention from both academic and industry.}
It has demonstrated impressive 
%results in various synthesis applications, 
\bihan{outcomes across diverse generative applications,}
\eg, text-to-image generation~\cite{adm,ldm,vqdm,sdxl}, image-to-image translation~\cite{saharia2022palette,su2022dual}, and text-to-video generation~\cite{lvdm,make-a-video,blattmann2023align,wu2023tune,yu2023video}.
%As this field continues to evolve, there is a growing demand for higher-quality generated images and videos.
%\bihan{One common challenge remains is how to efficiently scale up the trained SD to higher-quality generation.}
\bihan{To scale up SD models to high-resolution content generation, a commonly employed approach is progressive training ~\cite{sd2-1-base,sdxl}, \ie, training the SD model with lower-resolution images before fine-tuning with higher-resolution images.}
This warm-up approach enhances the model's semantic composition ability, \bihan{leading to the generation of high-quality images.} 
%and generates high-quality images. 
However, even a well-trained diffusion model for low-resolution images \bihan{demands extensive} fine-tuning and computational resources when transferring to a high-resolution domain due to its large size of model parameters. For instance, SD 2.1~\cite{sd2-1-base} requires $550k$ steps of training at resolution $256^2$ before fine-tuning with $>\!1000k$ steps at resolution $512^2$ to enable $512^2$ image synthesis.
%Without sufficient tuning steps, the model composition ability may be severely degraded, 
\bihan{Insufficient tuning steps may severely degrade the model's composition ability,}
resulting in issues such as pattern repetition, desaturation, and unreasonable object structures as in \cref{fig:intro}.

%A few 
\bihan{Several} tuning-free methods, such as those proposed in \cite{trainfree-variablesize} and ScaleCrafter~\cite{he2023scalecrafter}, attempted to seamlessly adapt the SD to higher-resolution image generation with 
%much fewer efforts.
\bihan{reduced efforts.}
In \cite{trainfree-variablesize}, the authors explored SD adaptation for variable-sized image generation 
%from the perspective of 
\bihan{using}
attention entropy, while ScaleCrafter~\cite{he2023scalecrafter} utilized dilated convolution to enlarge the receptive field of convolutional layers and adapt to new resolution generation.
However, these tuning-free solutions require careful adjustment of factors such as the dilated stride and injected step, 
%which may fail 
\bihan{potentially failing}
to account for the varied scales of object generation.
%Recently, some methods, 
\bihan{More recent methods,}
such as those proposed in \cite{any-size-diffusion}, have attempted to utilize LORA~\cite{hu2021lora} as additional parameters for fine-tuning. 
\bihan{However, this approach is not specifically designed for scale adaptation and still requires a substantial number of tuning steps.}
%which is not specifically designed for the scale adaptation problem and still requires huge of tuning steps.
\bihan{Other works~\cite{ho2022cascaded,wang2023lavie,zhang2023show} proposed to cascade the super-resolution mechanisms based on diffusion models for scale enhancement.}
%Alternatively, efforts have been made to cascade diffusion models~\cite{ho2022cascaded,wang2023lavie,zhang2023show} to cascade the super-resolution  for scale enhancement.
However, the use of extra super-resolution models necessitates a doubling of training parameters and limits the scale extension ability for a higher resolution.

In this paper, we present a novel self-cascade diffusion model that harnesses the rich knowledge gained from a well-trained low-resolution generation model to enable rapid adaptation to higher-resolution generation.
% In this paper, we introduce a new self-cascade diffusion model that leverages the extensive knowledge acquired from a well-trained low-resolution generation model. By cyclically re-utilizing the low-resolution model, we enable swift adaptation to higher resolution generation.
Our approach begins with the introduction of a tuning-free version, which utilizes a pivot replacement strategy to enforce the synthesis of detailed structures at a new scale by injecting reliable semantic guidance derived from the low-resolution model.
Building on this \bihan{baseline}, we further propose time-aware feature upsampling modules as plugins to a base low-resolution model to conduct a tuning version.
To enhance the robustness of scale adaptation while preserving the model's original composition and generation capabilities, we fine-tune the plug-and-play and lightweight upsampling modules at different feature levels, using a small amount of acquired high-quality data with a few tuning steps.

The proposed upsampler modules can be flexibly plugged into any pre-trained SD-based models, including both image and video generation models.
Compared to full fine-tuning, our approach offers a training speed-up of more than 5 times and requires only 0.002M trainable parameters. 
\bihan{Extensive experiments demonstrated that our proposed method can rapidly adapt to higher-resolution image and video synthesis with just $10k$ fine-tuning steps and virtually no additional inference time.}

Our main contributions are summarized as follows:
\begin{itemize}

    \item We propose a novel self-cascade diffusion model for fast-scale adaptation to higher resolution generation, by cyclically re-utilizing the low-resolution diffusion model. 
    Based on that, we employ a pivot replacement strategy to enable a tuning-free version as the baseline.

    \item We further construct a series of plug-and-play, learnable time-aware feature upsampler modules to incorporate knowledge from a few high-quality images for fine-tuning. This approach achieves a $5\times$ training speed-up compared to full fine-tuning and requires only 0.002M learnable parameters.

    \item Comprehensive experimental results on image and video synthesis demonstrate that the proposed method attains state-of-the-art performance in both tuning-free and tuning settings across various scale adaptations.
\end{itemize}

\section{Related Work}
\label{sec:relatedwork}

\noindent\textbf{Stable diffusion.}
\bihan{Building upon the highly effective and efficient foundations established by the Latent Diffusion Model (LDM)~\cite{rombach2022high},}
diffusion models~\cite{ho2020denoising,song2020score} have recently \bihan{demonstrated} remarkable performance in various practical applications, \eg, text-to-image generation~\cite{adm,ldm,vqdm,sdxl}, image-to-image translation~\cite{saharia2022palette,su2022dual}, and text-to-video generation~\cite{lvdm,make-a-video,blattmann2023align,wu2023tune,yu2023video}.
In this field, stable diffusion (SD)~\cite{rombach2022high,sdxl} has \bihan{emerged as a} prominent model for generating photo-realistic images from text.
However, despite its impressive synthesis capabilities at specific resolutions (\eg, $512^2$ for SD 2.1 and $1024^2$ for SD XL), it often produces extremely unnatural outputs for unseen image sizes. This limitation mainly arises from the fact that current SD models are trained exclusively on fixed-size images, leading to a lack of varied resolution generalizability. In this paper, we aim to explore the fast adaptation ability of the original diffusion model with limited image size to a higher resolution.

\vspace{1mm}
\noindent\textbf{High-resolution synthesis and adaptation.}
Although existing stable diffusion-based synthesis methods \bihan{have achieved} impressive results, 
\bihan{high-resolution image generation remains challenging and demands substantial computational resources,}
%high-resolution image synthesis is still challenging that requires significant computational resources, 
primarily due to the complexity of learning from higher-dimensional data.
%Besides, large-scale high-quality image and video training dataset is much more difficult to collect in practice, which also limits the synthesis performance.
\bihan{Additionally, the practical difficulty of collecting large-scale, high-quality image and video training datasets further constrains synthesis performance.}
To address \bihan{these challenges}, prior work can be \bihan{broadly categorized} into three \bihan{main approaches}:
\begin{enumerate}
\item Training from scratch. 
This type of work can be further divided into two categories: cascaded models~\cite{cascaded-dm,relay-diffusion,gu2023matryoshka,ho2022cascaded} and end-to-end models~\cite{simple-diffusion, importance-diffusion, sdxl, bond2023inftydiff}.
Cascade diffusion models employ an initial diffusion model to generate lower-resolution data, followed by a series of super-resolution diffusion models to successively upsample it. 
End-to-end methods learn a diffusion model and directly generate high-resolution images in one stage.
However, they all necessitate sequential, separate training and a significant amount of training data at high resolutions.

    \item Fine-tuning. Parameter-efficient tuning is an intuitive solution for higher-resolution adaptation. DiffFit~\cite{difffit} utilized a customized partial parameter tuning approach for general domain adaptation. Zheng~\etal~\cite{any-size-diffusion} adopted the LORA~\cite{hu2021lora} as the additional parameters for fine-tuning, which is still not specifically
designed for the scale adaptation problem and still requires
huge of tuning steps.
    \item Tuning-free. Several methods~\cite{trainfree-variablesize,he2023scalecrafter,du2024demofusion,haji2024elasticdiffusion} have explored expanding low-resolution diffusion models to higher resolutions without tuning. Recently, Jin~\etal\cite{trainfree-variablesize} explored a tuning-free approach for variable sizes but did not address high-resolution generation. 
    ScaleCrafter~\cite{he2023scalecrafter} employed dilated convolution to expand the receptive field of convolutional layers for adapting to new resolutions. 
    Besides, DemoFusion~\cite{du2024demofusion} used low-resolution semantic guidance as well as dilated sampling strategy to achieve a high-resolution generation.
    However, these approaches require careful adjustments, such as dilated stride and injected step, which lack semantic constraints and result in artifacts for various generation scales.
\end{enumerate}

\section{Preliminary}

\bihan{Our proposed method is based on the recent text-to-image diffusion model (\ie, stable diffusion (SD)~\cite{rombach2022high,sdxl}),}
which formulates the diffusion and denoising process in a learned low-dimensional latent space.
An autoencoder first conducts perceptual compression to significantly reduce the computational cost, where the encoder $E$ converts image $x_0 \in \mathbb{R}^{3\times H \times W}$ to its latent code $z_0 \in \mathbb{R}^{4 \times H' \times W'}$ and the decoder $D$ reconstructs the image $x_0$ from $z_0$ as follows,
\begin{equation}\label{eq:vae}
    z_0 = E(x_0)\;, \quad \hat{x
}_0 = D(z_0) \approx x_0 \;.
\end{equation}
Then, the diffusion model formulates a fixed forward diffusion process to gradually add noise to the latent code $z_0 \sim p(z_0)$:
\vspace{-1mm}
\begin{equation}\label{eq:forward}
    q(z_t|z_0) = \mathcal{N}(z_t; \sqrt{\bar{\alpha}_t} z_0, (1-\bar{\alpha}_t)\mathbf{I})\;.
\end{equation}
\vspace{-1mm}
In the inference stage, we sample latent features from the conditional distribution $p(z_0|c)$ with the conditional information $c$ (\eg, the text embedding extracted by a CLIP encoder~\cite{radford2021learning} $E_{CLIP}$):
\begin{equation}\label{eq:diffusion_reverse}
    p_\theta(z_{0:T}|c) =  p(z_T) \prod^{T}_{t=1}p_\theta(z_{t-1}|z_t, c).
\end{equation}
% \vspace{-1mm}
The U-Net denoiser $\epsilon_\theta$ consists of a sequential transformer and convolution blocks to perform denoising in the latent space.  
The corresponding optimization process can be defined as the following formulation:
\begin{equation}\label{diffusion_reverse}
    \mathcal{L} =  \mathbb{E}_{z_t, c, \epsilon, t}(\|\epsilon - \epsilon_\theta(z_t, t, c)\|^2),
\end{equation}
where $z_t = \sqrt{\bar{\alpha}_t} z_0+\sqrt{1-\bar{\alpha}_t} \epsilon$ represents the noised feature map at step $t$. 

\begin{figure*}[!t]
    \centering
    % \vspace{-0.3cm}
    \includegraphics[width=.9\linewidth]{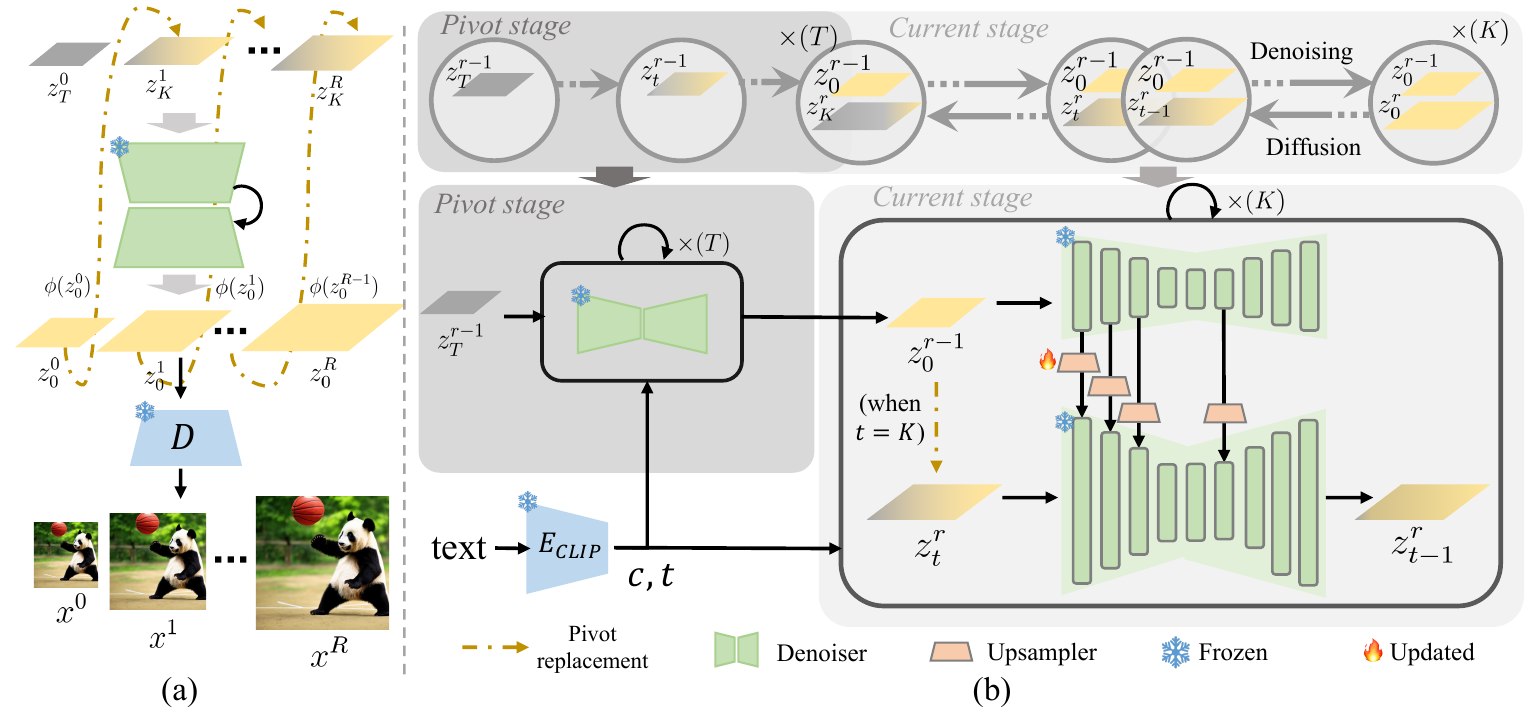}
    \vspace{-0.2cm}
    \caption{Illustration of the proposed self-cascade diffusion model, which is implemented in both tuning-free and tuning versions.
    (a) For the tuning-free version, we cyclically re-utilize the low-resolution model to progressively adapt it to the higher-resolution generation; 
    (b) For the tuning version, we additionally plug feature upsamplers ($\Phi$) into the base low-resolution generation model:
 the denoising process of image $z^r_t$ in step $t$ will be guided by the pivot guidance $z^{r-1}_0$ from the pivot stage (last stage) with a series of plugged-in tuneable upsampler modules.}
    \label{fig:framework}
    \vspace{-0.25cm}
\end{figure*}

\section{Methodology}
\label{sec:methodology}
In this section, we first introduce the overall framework of the proposed self-cascade diffusion model (Sec.~\ref{sec:problem}).
Based on that, we propose a tuning-free version using a pivot replacement strategy as the baseline, as well as an improved tuning version by plugging tunable feature upsamplers (Sec.~\ref{sec:upsampler}).
We then provide an analysis and discussion on our self-cascade diffusion model (Sec.~\ref{sec:analysis}).

\subsection{Self-Cascade Diffusion Model}\label{sec:problem}
Given a pre-trained stable diffusion (SD) model with the denoiser $\epsilon_\theta(\cdot)$ for synthesizing low-resolution images (latent code) $z\in \mathbb{R}^{d}$, our goal is to generate higher-resolution images $z^R\in \mathbb{R}^{d_R}$ in a time/resource and parameter-efficient manner with an adapted model $\tilde{\epsilon}_\theta(\cdot)$. 
To achieve such a goal, we aim to reuse the rich knowledge from the well-trained low-resolution model and only learn the low-level details at a new scale. 
Thus, we propose a self-cascade diffusion model to cyclically re-utilize the low-resolution image synthesis model. 
We intuitively define a \textbf{\textit{scale decomposition}} to decompose the whole scale adaptation $\mathbb{R}^d \rightarrow \mathbb{R}^{d_R}$ into multiple progressive adaptation processes such that $d=d_0<d_1 \ldots <d_R$ where $R = \left \lceil \text{log}_4{ d_R / d}\right \rceil$. 
We first progressively synthesize a low-resolution image (latent code) $z^{r-1}$ and then utilize it as the pivot guidance to synthesize the higher resolution result $z^{r}$ in the next stage, where the reverse process of the self-cascade diffusion model can be extended by Eq. (\ref{eq:diffusion_reverse}) for each $z^r$, $r=1, \ldots, R$ as follows:
% \vspace{-1mm}
\begin{equation}\label{diffusion_reverse}
    p_\theta(z^r_{0:T}|c, z^{r-1}_0) =  p(z^r_T)\prod^{T}_{t=1} p_\theta(z^r_{t-1}|z^r_t, c, z^{r-1}_0),
\end{equation}
% \vspace{-1mm}
where the reverse transition $p_\theta(z^r_{t-1}|z^r_t, c, z^{r-1}_0)$ not only conditions on denoising step $t$ and text embedding $c$, but also on lower-resolution latent code $z^{r-1}_0$ generated in the last stage.
Different from previous works, \eg, \cite{cascaded-dm}, LAVIE~\cite{wang2023lavie}, and SHOW-1~\cite{zhang2023show} where $p_{\theta}$ in Eq.~\ref{diffusion_reverse} is implemented by a new super-resolution model, we cyclically re-utilize the base low-resolution synthesis model to inherit the prior knowledge of the base model thus improve the efficiency.

\vspace{1mm}
\noindent\textbf{Pivot replacement.}\label{sec:pivot}
According to the \textit{scale decomposition}, the whole scale adaptation process will be decoupled into multiple moderate adaptation stages, \eg, $4\times$ more pixels than the previous stage.
The information capacity gap between $z^r$ and $z^{r-1}$ is not significant, especially in the presence of noise (intermediate step of diffusion).
Consequently, we assume that $p(z_K^{r}|z_0^{r-1})$ can be considered as the proxy for $p(z_K^{r}|z_0^{r})$ to manually set the initial diffusion state for current adaptation stage $\mathbb{R}^{d_{r-1}} \rightarrow \mathbb{R}^{d_r}$, where $K<T$ is an intermediate step.
Specifically, let $\phi$ denote a deterministic resize interpolation function (\ie, bilinear interpolation) to upsample from scale $d_{r-1}$ to $d_r$. We upsample the generated lower-resolution image $z_0^{r-1}$ from last stage into $\phi(z_0^{r-1})$ to maintain dimensionality.
Then we can diffuse it by $K$ steps and use it to replace $z_K^{r}$ as follows:
\begin{equation}\label{diffusion_reverse}
    z_K^r \sim \mathcal{N}(\sqrt{\bar{\alpha}_K}\phi(z_0^{r-1}), \sqrt{1-\bar{\alpha}_K} \mathbf{I}).
\end{equation}
% \vspace{-1mm}
Regarding $z_K^r$ as the initial state for the current stage and conduct denoising with $K \rightarrow 0$ steps as Eq. (\ref{eq:diffusion_reverse}) to generate the $z_0^r$, which is the generated higher-resolution image in the current stage.

We can employ such a pivot replacement strategy at all decoupled scale adaptation stages.
Hence, the whole synthesis process for a higher-resolution image with resolution $d_R$ using pivot replacement strategy can be illustrated as \cref{fig:framework}(a).
So far, we have devised a \textbf{\textit{tuning-free version}} of self-cascade diffusion model (denoted as Ours-TF in experiments) to progressively expand the model capacity for higher-resolution adaptation with cyclically re-utilizing the totally frozen low-resolution model.
Although the tuning-free self-cascade diffusion model built upon pivot replacement strategy (Sec.~\ref{sec:pivot}) can achieve a feasible and scale-free higher-resolution adaptation, 
it has limitations on synthesis performance especially the detailed low-level structures due to the unseen higher-resolution ground-truth images.
To achieve a more practical and robust scale adaptation performance, we further introduce an improved \textbf{\textit{tuning version}} of the self-cascade diffusion model (denoted as Ours-T in experiments) in Sec.~\ref{sec:upsampler}.

\subsection{Feature Upsampler Tuning}\label{sec:upsampler}

In this section, we propose a tuning version of the self-cascade diffusion model that enables a cheap scaling, by inserting very lightweight time-aware feature upsamplers as illustrated in \cref{fig:framework}(b). The proposed upsamplers can be plugged into any diffusion-based synthesis methods. 
The detailed tuning and inference processes of our tuning version self-cascade diffusion model are in Algorithm~\ref{algo_train} and~\ref{algo}, respectively. Note that by omitting the tuning process and solely executing the inference step in Algorithm~\ref{algo}, it turns into our tuning-free version.

Specifically, given an intermediate $z^r_{t}$ in step $t$ and the pivot guidance $z^{r-1}_0$ from the last stage, we can achieve corresponding intermediate multi-scale feature groups $h_t^r$ and $h_0^{r-1}$ via the pre-trained UNet denoiser $\epsilon_\theta$, respectively, as follows:
\begin{align}\label{diffusion_reverse}
   \nonumber h^{r-1}_0 =& \{h^{r-1}_{1,0}, h^{r-1}_{2,0}, \ldots, h^{r-1}_{N,0}\} \\ 
       h^{r}_t =& \{h^{r}_{1,t}, h^{r}_{2,t}, \ldots, h^{r}_{N,t}\} 
\end{align}
where $N$ represents the number of features within each feature group and the details are included in the \textbf{supplementary}.
In short, inspired by the recent work~\cite{si2023freeu} that investigated the impact of various components in the UNet architecture on synthesis performance, we choose to use skip features as a feature group. 
These features have a negligible effect on the quality of the generated images while still providing semantic guidance. We define a series of time-aware feature upsamplers $\Phi=\{\phi_1, \phi_2, \ldots, \phi_N\}$ to upsample and transform pivot features at each corresponding scale.
During the diffusion generation process, the focus shifts from high-level semantics to low-level detailed structures as the signal-to-noise ratio progressively increases as noise is gradually removed. Consequently, we propose that the learned upsampler transformation should be adaptive to different time steps. 
The upsampled features $\phi_n(h^{r-1}_{n,0}, t)$ is then added to original features $h^{r}_{n,t}$ at each scale:
\begin{align}
\hat{h}^{r}_{n,t} = h^{r}_{n,t} + \phi_n(h^{r-1}_{n,0}, t), \;\; n\in\{1,\ldots, N\}.
\end{align}

\noindent \textbf{Optimization details.}
For each training iteration for scale adaptation $\mathbb{R}^{d_{r-1}} \rightarrow \mathbb{R}^{d_r}$, we first randomly sample a step index $t \in (0, K]$.
The corresponding optimization process can be defined as the following formulation:
\begin{equation}\label{diffusion_reverse}
    \mathcal{L} =  \mathbb{E}_{z^r_t, z^{r-1}_0, t, c, \epsilon, t}(\|\epsilon - \tilde{\epsilon}_{\theta,\theta_{\Phi}}(z_t^r, t, c, z^{r-1}_0)\|^2),
\end{equation}
where $\theta_{\Phi}$ denotes the trainable parameters of the plugged-in upsamplers and $\theta$ denotes the frozen parameters of pre-trained diffusion denoiser.
Each upsampler is simple and lightweight, consisting of one bilinear upsampling operation and two residual blocks. In all experiments, we set $N=4$, resulting in a total of 0.002M trainable parameters. Therefore, the proposed tuning self-cascade diffusion model requires only a few tuning steps (\eg, $10k$) and the collection of a small amount of higher-resolution new data.

\subsection{Analysis and Discussion}\label{sec:analysis}
% \noindent\textbf{Why it works?}
Drawing inspiration from previous explorations on scale adaptation~\cite{he2023scalecrafter}, we found that directly applying the SD 2.1 model trained with $512^2$  images to generate $1024^2$ images led to issues such as object repetition and diminished composition capacity (see \cref{fig:intro}). We observed that the local structural details of the generated images appeared reasonable and abundant without smoothness when the adapted scale was not large (\eg, $4\times$ more pixels). 
In summary, the bottleneck for adapting to higher resolutions lies in the semantic component and composition capacity. 
Fortunately, the original pre-trained low-resolution diffusion model can generate a reliable low-resolution pivot, naturally providing proper semantic guidance by injecting the pivot semantic features during the higher-resolution diffusive sampling process.
Simultaneously, the local structures can be completed based on the rich texture prior learned by the diffusion model itself, under strong semantic constraints.

\begin{figure}[t]
\centering
\vspace{-5mm}
\begin{minipage}[t]{0.45\linewidth}
\begin{algorithm}[H]\small
% \hspace*{\algorithmicindent}\noindent \textbf{Input:} natural image $\mathbf{x}$, shadow-free image $\mathbf{x}$, and initial mask $\mathbf{\tilde{m}}$.
\centering
\caption{Feature upsampler tuning process.}\label{algo_train}
\begin{algorithmic}[1]
%\Require Frozen parameter $\theta$ of the base model, tunable parameter $\theta_\Phi$
        \While{not converged}
        \State $(x_0, c) \sim p(x_0, c)$
        \State $z^{r}_0 = E(x_0)$
        \State $z^{r-1}_0 = E(Downsample(x_0))$
        \State $t \sim \text{Uniform}\{1, \ldots, K\}$
        \State $\epsilon \sim \mathcal{N}(\mathbf{0},\mathbf{I})$
        \State $z_t^r = \sqrt{\bar{\alpha}_t} z_0^r + \sqrt{1 - \bar{\alpha}_t} \epsilon$
        \State $\theta_\Phi \leftarrow \theta_\Phi - \eta \bigtriangledown_{\theta_\Phi} \| \tilde{\epsilon}_{\theta,\theta_{\Phi}}(z_t^r, t, c, z_0^{r-1}) - \epsilon\|^2$
        \EndWhile
      \State $\textbf{return}\; \theta_{\Phi}$
\end{algorithmic}
\end{algorithm}
\end{minipage}
% \hfill
\hspace{2mm}
\begin{minipage}[t]{0.45\linewidth}
\centering
% \vspace{-0.2cm}
\begin{algorithm}[H]\small
\caption{Inference process for $\mathbb{R}^{d_{r-1}} \rightarrow \mathbb{R}^{d_{r}}$.}\label{algo}
\begin{algorithmic}[1]
\Require text embedding $c$
\If{$r=1$}
\State $z_T^r \sim \mathcal{N}(0, \mathbf{I})$

      \For{$t =T, \ldots, 1$}
        % \State $t = (s-1)\cdot T / S +1$
        % \State  $t_\text{next} = (s-2)\cdot T / S +1$ \;\textbf{if} $s>1$ \textbf{else} $0$
        \State $z^{r}_{t-1} \sim p_\theta(z^{r}_{t-1}|z^{r}_t, c)$
      \EndFor
      \Else
       \State $z^{r}_{K} \sim q(z^{r}_{K} | z^{r-1}_0)$
      \For{$t =K, \ldots, 1$}
        % \State $t = (s-1)\cdot T / S +1$
        % \State  $t_\text{next} = (s-2)\cdot T / S +1$ \;\textbf{if} $s>1$ \textbf{else} $0$
        \State $z^{r}_{t-1} \sim p_\theta(z^{r}_{t-1}|z^{r}_t, c, z^{r-1}_0)$
      \EndFor
      \EndIf
 
      \State \textbf{return} $z^{r}_0$
\end{algorithmic}
\end{algorithm}
\end{minipage}
\vspace{-2mm}
\end{figure}

% \noindent\textbf{Relation to other cascade methods:}
Compared to existing cascaded diffusion frameworks for high-fidelity image and video generation~\cite{cascaded-dm}, our work is the first to conduct self-cascade by cyclically re-utilizing pre-trained diffusion model on low-resolution with the following major advantages:
\begin{itemize}
\item \textbf{Lightweight upsampler module.} Conventional cascade diffusion models comprise a pipeline of multiple diffusion models that generate images of increasing resolution, 
which results in a multiplicative increase in the number of model parameters. Our model is built upon the shared diffusion model at each stage with only very lightweight upsampler modules (\ie, 0.002M parameters) to be tuned.

\item \textbf{Less fine-tuning data.}
Previous cascaded model chains necessitate sequential, separate training, with each model being trained from scratch, thereby imposing a significant training burden. Our model is designed to quickly adapt the low-resolution synthesis model to higher resolutions using a small amount of high-quality data for fine-tuning.
\end{itemize}

\section{Experiments}
\label{sec:experiments}

\begin{table*}[t]
    \small
        \centering
            \caption{Quantitative results of different methods on the dataset of \textit{Laion-5B} with $4\times$ adaptation on $1024^2$ resolution. The best results are highlighted in \textbf{bold}. Note that Ours-TF and Ours-T denote the tuning-free version and the upsampler tuning version, respectively. ``\#Param'' denotes the number of trainable parameters and ``Infer Time'' denotes the inference time of different methods \textit{v.s.} Direct Inference. 
            pFID$_r$/pKID$_r$ denote the patch-
            We put `-' since FID$_b$/KID$_b$ are unavailable for SD+SR\protect\footnotemark.}
       % \vspace{-2mm}
    \scalebox{.8}{
   \begin{tabular}{c|c|c|c|cccccc}
        \Xhline{0.8pt}
        Methods & \#Param & Training Step & Infer Time & FID$_r$$\downarrow$ & KID$_r$$\downarrow$ & 
        pFID$_r$$\downarrow$ & pKID$_r$$\downarrow$ &
        FID$_b$$\downarrow$ & KID$_b$$\downarrow$ \\
            \Xhline{0.4pt}
            Direct Inference  & 0 & - & $1\times$ &29.89 & 0.010 & 20.88& \textbf{0.0070} & 24.21 & 0.007   \\
            Attn-SF~\cite{trainfree-variablesize}  & 0  & - & $1\times$ & 29.95 & 0.010 &21.07 &0.0072 &  22.75 & 0.007  \\
            ScaleCrafter~\cite{he2023scalecrafter} & 0 & -&  $1\times$ & 20.88 & 0.008 & 21.00 & 0.0071& 16.67 & 0.005  \\
            \textbf{Ours-TF} (\textbf{T}uning-\textbf{F}ree)   & 0 & - & $1.04\times$ & \textbf{12.25} & \textbf{0.004} & \textbf{19.59} & 0.0071 &\textbf{6.09} & \textbf{0.001}  \\
            \hline
            Full Fine-tuning    & 860M & $18k$ &  $1\times$ & 21.88 & 0.007 & 19.33 & 0.0077& 17.14 & 0.005 \\
            % Full Fine-tuning ($100k$)   &  & $1\times$ &   \\
            LORA-R32   & 15M  & $18k$&$1.22\times$ &  17.02 & 0.005 & 18.65& 0.0076 & 11.33 & 0.003 \\
            LORA-R4   & 1.9M &$18k$ &$1.20\times$ & 14.74 & 0.005 & 18.06 & 0.0074 & 9.47 & 0.002 \\
            SD+SR & 184M & 1.25M & $5\times$ & 12.59 & 0.005 &17.21&\textbf{0.0053}  & -  & - \\
            \textbf{Ours-T} (\textbf{T}uning)  & 0.002M &  $4k$ &$1.06\times$ &  \textbf{12.40} & \textbf{0.004} & \textbf{15.35} & 0.0058& \textbf{3.15} & \textbf{0.0005}\\
       \Xhline{0.8pt}
    \end{tabular} 
    }
    % \vspace{-2mm}
    \label{tab:image_4}
    % \vspace{-2mm}
\end{table*}

\footnotetext{We follow the same comparison settings of ScaleCrafter~\cite{he2023scalecrafter}. Since FID$_b$/KID$_b$ are evaluated on the original low-resolution by down-sampling, the down-sampling results of SD+SR will be roughly the same as the reference real image set which denotes ``zero distance''.}

\begin{table*}[!t]
    \centering
        \caption{Quantitative results of different methods on the dataset of \textit{Laion-5B} with $16\times$ image scale adaptation to $2048^2$ resolution. The best results are highlighted in \textbf{bold}. $10k$ and $20k$ denote the training steps used for tuning.}
     % \vspace{-2mm}
    \scalebox{0.8}{
    \begin{tabular}{c|cccc}
    \toprule
   Methods &FID$_r$$\downarrow$& KID$_r$$\downarrow$& FID$_b$$\downarrow$& KID$_b$$\downarrow$  \\
    \midrule
    Direct Inference & 104.70 & 0.043 & 104.10 & 0.040  \\
    Attn-SF~\cite{trainfree-variablesize} & 104.34 & 0.043 & 103.61 & 0.041\\
     ScaleCrafter~\cite{he2023scalecrafter} & 59.40 & 0.021 & 57.26 & 0.018 \\
     \textbf{Ours-TF} (\textbf{T}uning-\textbf{F}ree) &  \textbf{38.99} & \textbf{0.015} & \textbf{34.73} & \textbf{0.013} \\
     \midrule
    Full Fine-tuning ($20k$)  & 43.55 &0.014   &  41.58 & 0.012 \\
    LORA-R4 ($20k$) & 50.72 & 0.020 &51.99 & 0.019 \\
     % Full Fine-tuning & 531.57 & 33.61 \\
    \textbf{Ours-T} (\textbf{T}uning) ($10k$) &\textbf{18.46}&\textbf{0.005}& \textbf{8.99} &\textbf{0.001}  \\
    \bottomrule
    \end{tabular}}
    % \vspace{-2mm}
    \label{tab:image_16}
\end{table*}

\subsection{Implementation Details}
The proposed method is implemented using PyTorch and trained on two NVIDIA A100 GPUs. The original base diffusion model's parameters are frozen, with the only trainable component being the integrated upsampling modules.
The initial learning rate is $5\times 10^{-5}$.
We used $1000$ diffusion steps $T$ for training, and $50$ steps for DDIM~\cite{ddim} inference.
We set $N=4$ and $K=700$ for all experiments.
We conduct evaluation experiments on text-to-image models, specifically Stable Diffusion (SD), focusing on two widely-used versions: SD 2.1~\cite{sd2-1-base} and SD XL 1.0~\cite{sdxl}, as they adapt to two unseen higher-resolution domains. For the original SD 2.1, which is trained with $512^2$ images, the inference resolutions are $1024^2$ and $2048^2$, corresponding to $4\times$ and $16\times$ more pixels than the training, respectively.
We also conduct evaluation experiments on text-to-video models, where we select the LVDM~\cite{lvdm} as the base model which is trained with $16\times 256^2$ videos ($16$ frames), the inference resolutions are $16\times 512^2$, $4\times$ more pixels than the base resolution.
We left the experiments for SD XL 1.0 in the \textbf{supplementary}. 

\subsection{Evaluation on Image Generation}

% \noindent\textbf{Experimental setup.} 

% \vspace{1mm}
\noindent\textbf{Dataset and evaluation metrics.}
We select the Laion-5B~\cite{laion5b} as the benchmark dataset which contains 5 billion image-caption pairs. 
We fine-tune all tuning-based competing methods by applying online filtering on Laion-5B for high-resolution images larger than the target resolution.
We randomly sample $30k$ images with text prompts from the dataset and evaluate the generated image quality and diversity using the Inception Distance (FID) and Kernel Inception Distance (KID) metrics, which are measured between the generated images and real images, denoted as FID$_r$ and KID$_r$. 
Following previous work~\cite{he2023scalecrafter}, we sample $10k$ images when the inference resolution is higher than $1024^2$.
Besides, to address the issue of squeezed resolutions in standard FID$_r$/KID$_r$, we randomly cropped local patches to calculate these metrics instead of resizing, denoted as pFID$_r$/pKID$_r$~\cite{chai2022any}.
To ensure consistency in image pre-processing steps, we use the clean-fid implementation~\cite{clean-fid}.
Since pre-trained models can combine different concepts that are not present in the training set, we also measure the FID and KID metrics between the generated samples under the base training resolution and inference resolution, denoted as FID$_b$ and KID$_b$. 
This evaluation assesses how well our method preserves the model's original ability when adapting to a new higher resolution.

% We sample 30K images with randomly sampled text prompts from the dataset.
% Following the standard evaluation protocol, we measure the Inception Distance (FID) and Kernel Inception Distance (KID) between generated images and real images to evaluate the generated image quality and diversity, referred to as FID$_r$ and KID$_r$.
% We adopt the implementation of clean-fid～\cite{} to avoid discrepancies in the image pre-processing steps. Since the pre-trained models have the capability of compositing different concepts that do not appear in the training set, we also measure the metrics between the generated samples under the base training resolution and inference resolution, referred to as FID$_b$ and KID$_b$. This evaluates how well our method can preserve the model's original ability when sampling under a new resolution.

\begin{figure*}[!t]
    \centering
    % \vspace{-0.6cm}
    \includegraphics[width=1.\linewidth]{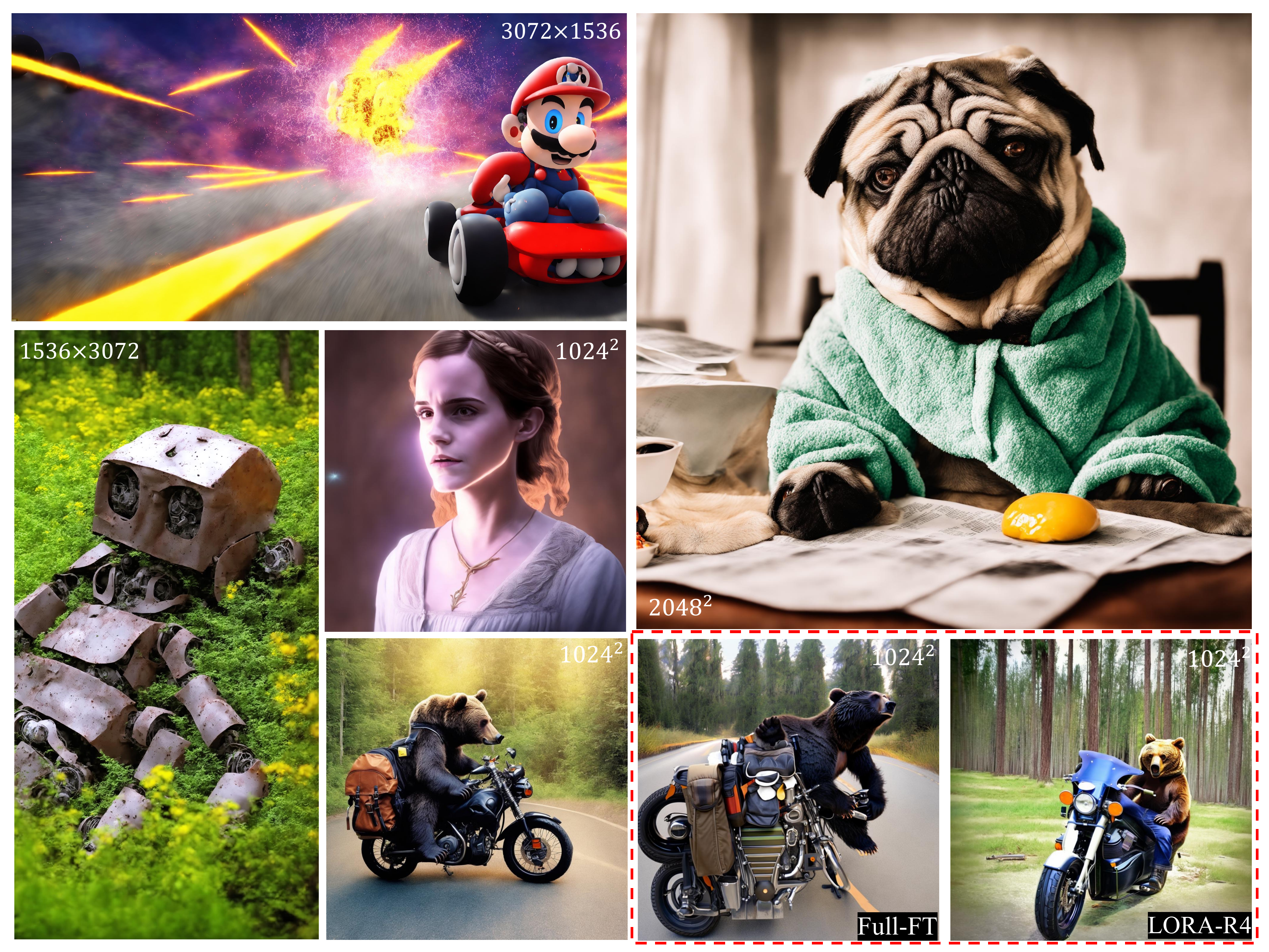}
    \vspace{-0.3cm}
    \caption{Visual examples of Ours-T (Tuning) on the higher-resolution adaptation to various higher resolutions, \eg, $1024^2$, $3072\times 1536$, $1536\times 3072$, and $2048^2$, with the pre-trained SD 2.1 trained with $512^2$ images, comparing to $1024^2$ results of Full fine-tuning (Full-FT) and LORA-R4 (right down corner: red dashed box). Please zoom in for more details.}
    \label{fig:image_4and16}
    \vspace{-0.25cm}
\end{figure*}

\vspace{1mm}
\noindent\textbf{Comparison with state-of-the-art.}
We conduct the comparison experiments on two settings, \ie, tuning-free and fine-tuning.
For the tuning-free setting, we compare our tuning-free version, denoted as Ours-TF, with the vanilla text-to-image diffusion model (Direct Inference) that directly samples the higher resolution images via the original checkpoint, as well as two tuning-free methods, \ie, Attn-SF~\cite{trainfree-variablesize} and ScaleCrafter~\cite{he2023scalecrafter}.
Besides, we also compare our fine-tuning version, denoted as Ours-T, with the full fine-tuning model, and LORA tuning (consisting of two different ranks, \ie, 4 and 32, denoted as LORA-R4 and LORA-R32).
%, and Any-Size-Diffusion~\cite{any-size-diffusion}.
% To better verify the effectiveness of scale adaptation, we also compare our method with using a super-resolution approach as post-processing, denoted as SD+SR.
\cref{tab:image_4} and \cref{tab:image_16} show the quantitative results on Laion-5B~\cite{laion5b} over $4\times$ and $16\times$ more pixels compared to base model.
Our methods outperform existing methods in both tuning-free and fine-tuning settings, especially when adapting to a higher-resolution domain, \eg, $16\times$ scale adaptation.
Besides, with the merits of injecting newly acquired higher-resolution data for tuning, our tuning version can achieve a better and more robust generation performance, especially for $16\times$ scale adaptation.
We show random samples from our method (Ours-T) on adapted various resolutions in \cref{fig:image_4and16}, comparing to results of Full fine-tuning (Full-FT) and LORA-R4.
Our method can achieve not only high-quality image results with rich structural details and accurate object composition, \eg, the relationship between the bear and motorbike as shown in \cref{fig:image_4and16}.
Visual comparisons with the competing methods are included in the \textbf{supplementary}.

\subsection{Evaluation on Video Generation}

\noindent\textbf{Dataset and evaluation metrics.}
We select the Webvid-10M~\cite{webvid} as the benchmark dataset which contains 10M high-resolution collected videos.
We randomly sample 2048 videos with text prompts from the dataset and evaluate the generated video quality and diversity using video counterpart Frechet Video Distance (FVD)~\cite{FVD} and Kernel Video Distance (KVD)~\cite{kvd}, denoted as FVD$_r$ and KVD$_r$.

\begin{figure*}[!t]
    \centering
    % \vspace{-0.3cm}
    \includegraphics[width=1.\linewidth]{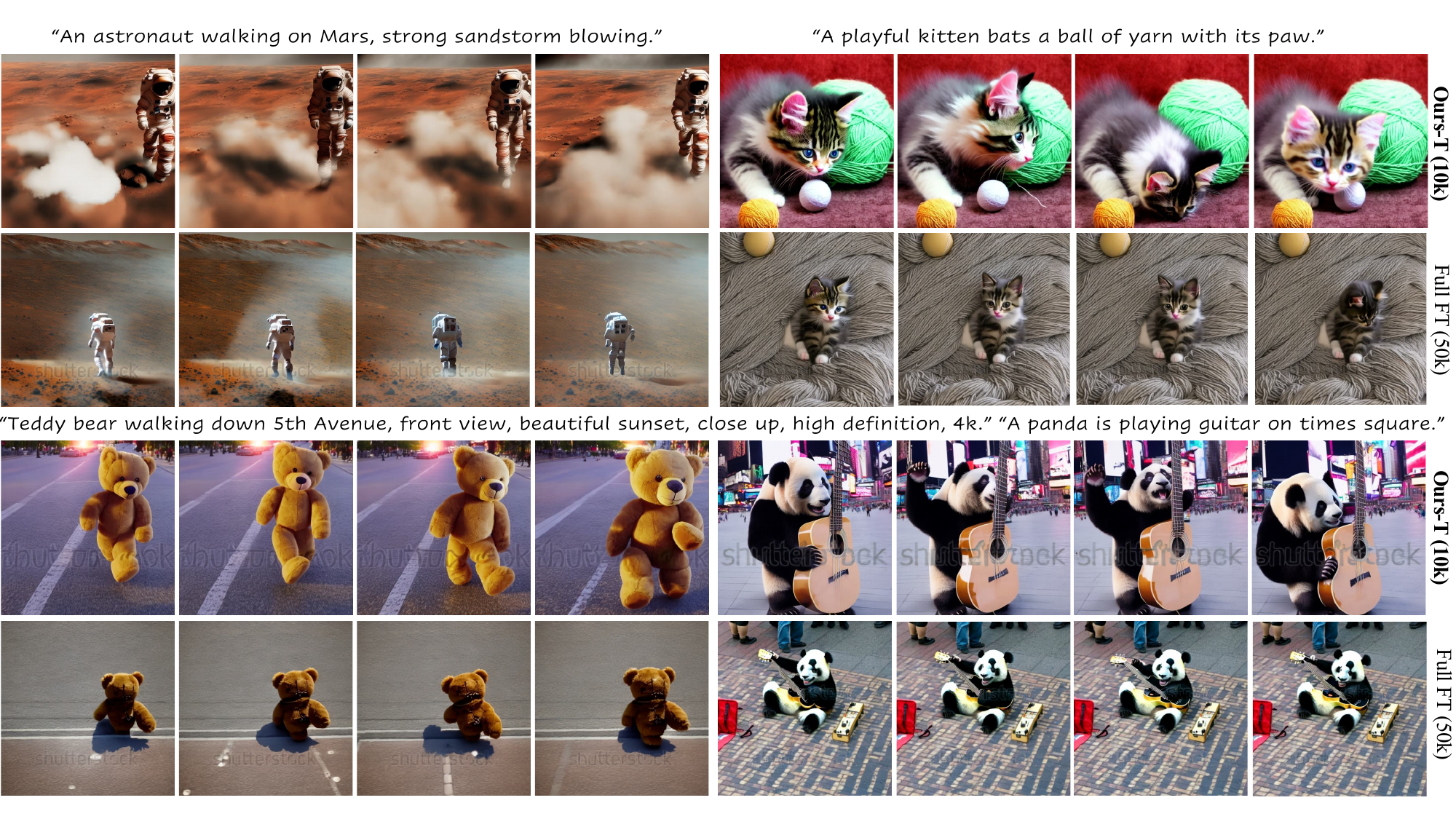}
    \vspace{-0.7cm}
    \caption{Visual quality comparisons between full fine-tuning ($50k$) and Ours-T ($10k$) on higher-resolution video synthesis of $16\times 512^2$. }
    \label{fig:video_4}
    % \vspace{-0.4cm}
\end{figure*}

\vspace{1mm}
\noindent\textbf{Comparison with state-of-the-art.}
To comprehensively verify the effectiveness of our proposed method, we also conduct comparison experiments on a video generation base model~\cite{lvdm}.
Thus, we compare our method with a full fine-tuning model and LORA tuning with different ranks, as well as the previous tuning-free method, \ie, ScaleCrafter. 
\cref{tab:video} shows the quantitative results on \textit{Webvid-10M}~\cite{webvid} and visual comparisons are shown in \cref{fig:video_4}.
Our method achieves better FVD and KVD results in approximately $20\%$ of the training steps compared to the competing approaches. 
With the merits of the reuse of reliable semantic guidance from a well-trained low-resolution diffusion model, our method can achieve better object composition ability (\eg, the reaction between cat and yarn ball and the times square as shown in the second and fourth examples of \cref{fig:video_4}, respectively) and rich local structures compared to the competing methods (\eg, the details of the teddy bear as shown in the third example of \cref{fig:video_4}).
In contrast, for full fine-tuning models, the issue of low saturation and over-smoothness requires many training steps to resolve and it is difficult to achieve results as good as those obtained with low-resolution models.
Besides, the generated results of both full fine-tuning and LORA tuning methods will have motion shift or motion inconsistency issues as shown in the bag of the astronaut in the first example of \cref{fig:video_4}, while our method can better maintain the original model's temporal consistency capabilities, generating more coherent videos (video examples refer to \textbf{supplementary}).

\subsection{Network Analysis}
\begin{figure}[!t]
    \centering
    \begin{minipage}{.55\textwidth}
        \centering
	\begin{center}
		\begin{tabular}{c@{ }c}
  % \vspace{-1mm}
			\includegraphics[width=.48\linewidth,height=.26\linewidth,trim={0.2cm 0.2cm 0.2cm 0.2cm},clip]{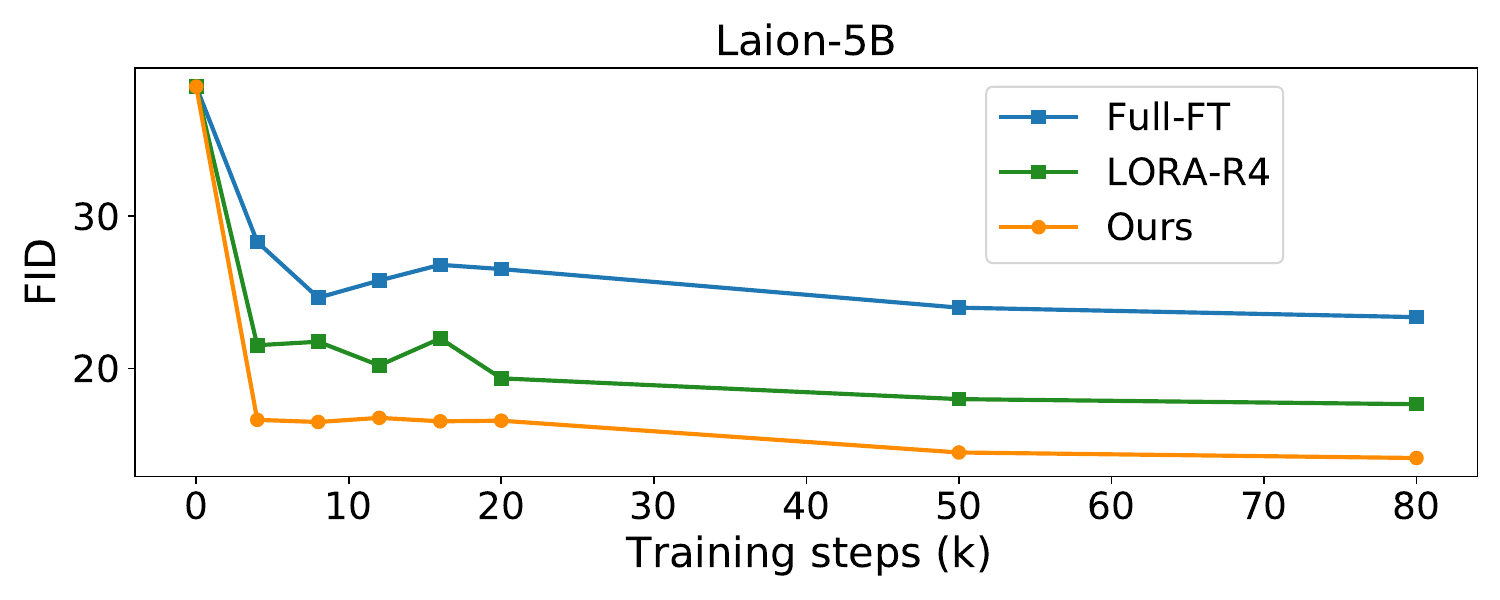}~&
			\includegraphics[width=.48\linewidth,height=.26\linewidth, trim={0.2cm 0.2cm 0cm 0.2cm},clip]{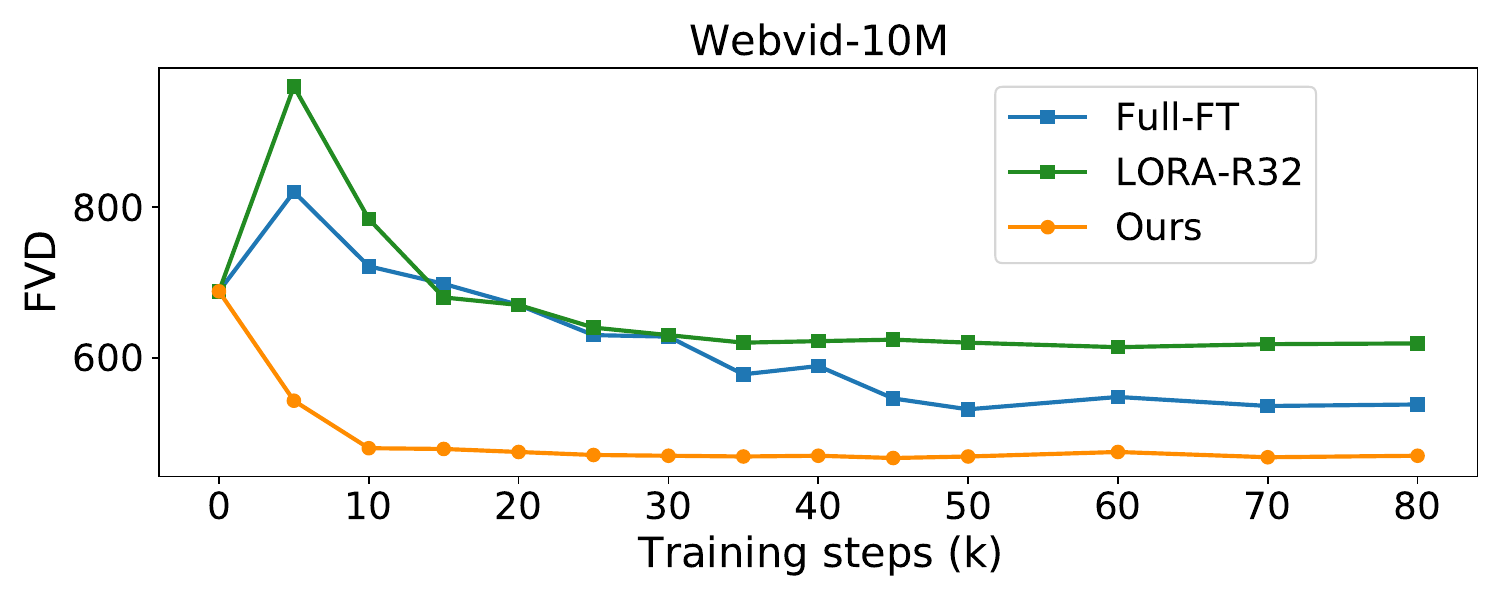}\\
		\end{tabular}
	\end{center}
 \vspace{-5mm}
 \caption{Average FID and FVD scores of three methods every $5k$ iterations on image (Laion-5B) and video (Webvid-10M) datasets. Our observations indicate that our method can rapidly adapt to the higher-resolution domain while maintaining a robust performance among both image and video generation.}
        \label{fig:finetune}
    \end{minipage}%
    \hfill
    \begin{minipage}{0.4\textwidth}
 	\begin{center}
   \vspace{-1mm}
			\includegraphics[width=1.\linewidth]{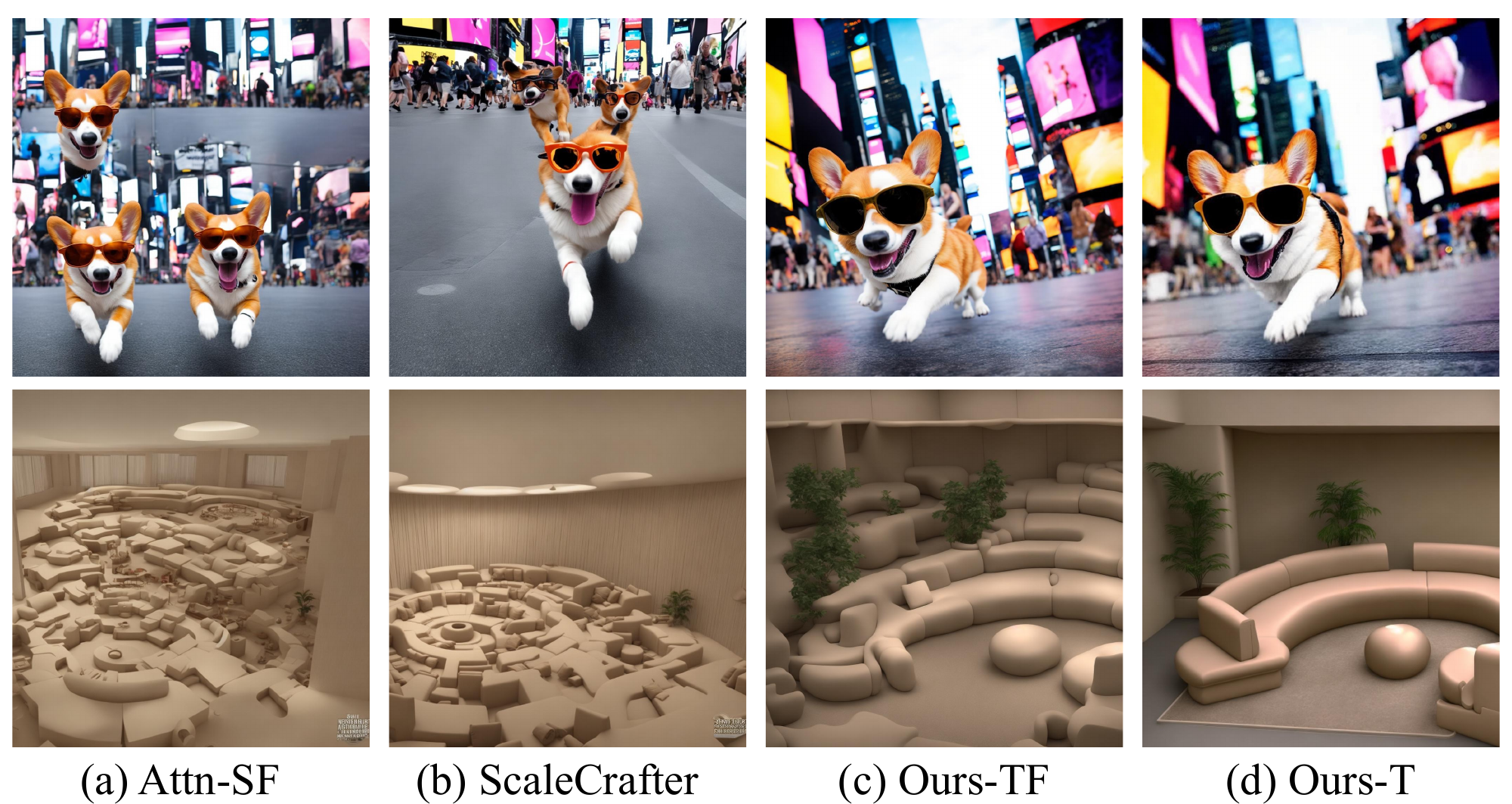}
	\end{center}
 \vspace{-6mm}
 \caption{Visual quality comparisons between the tuning-free methods and ours on higher-resolution adaptation to $1024^2$ resolutions. Please zoom-in to see more details.}
 \vspace{-0.4cm}
  \label{fig:ablation1}
    \end{minipage}
\end{figure}

\noindent\textbf{Efficiency comparison.}
To demonstrate the training and sampling efficiency of our method, we compare our approach with selected competing methods in \cref{tab:image_4} for generating $1024^2$ resolution images on the \textit{Laion-5B} dataset. Our model has only 0.002M trainable parameters, utilizing approximately the parameters compared to LORA-R4 (with a rank of 4). 
Although our proposed method requires a cascaded generation process, \ie, starting with low-resolution generation followed by progressively pivot-guided higher-resolution generation, the inference time of our method is similar to that of direct inference
(with a factor of $1.04 \times$ for the tuning-free version and $1.06\times$ for the tuning version), resulting in virtually no additional sampling time.
Besides, we present the FID and FVD scores for several methods every $5k$ iteration on image (Laion-5B) and video (Webvid-10M) datasets as shown in \cref{fig:finetune}.
Our observations demonstrate that our method can rapidly adapt to the desired higher resolution. By cyclically reusing the frozen diffusion base model and incorporating only lightweight upsampler modules, our approach maximally retains the generation capacity of the pretrained base model, resulting in improved fine-tuned performance.

% \noindent\textbf{Fine-tuning convergence analysis.}

% To demonstrate the training and sampling efficiency of our method, we compare our approach with the selected competing methods in Table~\ref{} on the Laion-5B dataset with resolution $1024^2$ images generation.
% Our model has only 0.002M trainable parameters, utilizing approximately $0.1\%$ of the parameters compared to LORA-R4 (with a rank of 4).
% While the proposed method requires the cascaded generation process, \ie, beginning with a low resolution generation, followed by progressively pivot guided higher-resolution generation, the inference time of our method is similar with ($1.04 \times$ for tuning-free version and $1.06\times$ for tuning version) the original baseline with virtually no extra sampling time.

% \subsection{Ablation Studies}

 \begin{table}[t]
\begin{minipage}[b]{0.4\linewidth}
\centering
%\vspace{-14mm}
\vspace{-4mm}
    \caption{Quantitative results of different methods on the dataset of \textit{Webvid-10M}~\cite{webvid} with $4\times$ video scale adaptation on $16\times 512^2$ resolution (16 frames). The best results are highlighted in \textbf{bold}. $10k$ and $50k$ denote the training steps used for tuning.}
    \vspace{-2mm}
        \renewcommand{\arraystretch}{0.8}
 \scalebox{.8}{
    \begin{tabular}{c|cc}
    \toprule
    Methods & FVD$_r$$\downarrow$& KVD$_r$$\downarrow$ \\
    \midrule
    Direct Inference & 688.07 & 67.17 \\
     ScaleCrafter~\cite{he2023scalecrafter} & 562.00 & 44.52 \\
     \textbf{Ours-TF} & \textbf{553.85} & \textbf{33.83}\\
     \midrule
    Full Fine-tuning ($10k$) & 721.32 & 94.57  \\
     Full Fine-tuning ($50k$) & 531.57 & 33.61 \\
     LORA-R4 ($10k$)&1221.46 &263.62\\
     LORA-R32 ($10k$) & 959.68 & 113.07\\
        LORA-R4 ($50k$) & 623.72 & 74.13\\
     LORA-R32 ($50k$) & 615.75 & 76.99\\
    \textbf{Ours-T} ($10k$) &  \textbf{494.19} & \textbf{31.55} \\
    \bottomrule
    \end{tabular}}
    \vspace{2mm}
    \label{tab:video}
\end{minipage}\hfill
\begin{minipage}[b]{0.55\linewidth}
\centering
	\vspace{3mm}		
   \includegraphics[width=1.\linewidth]{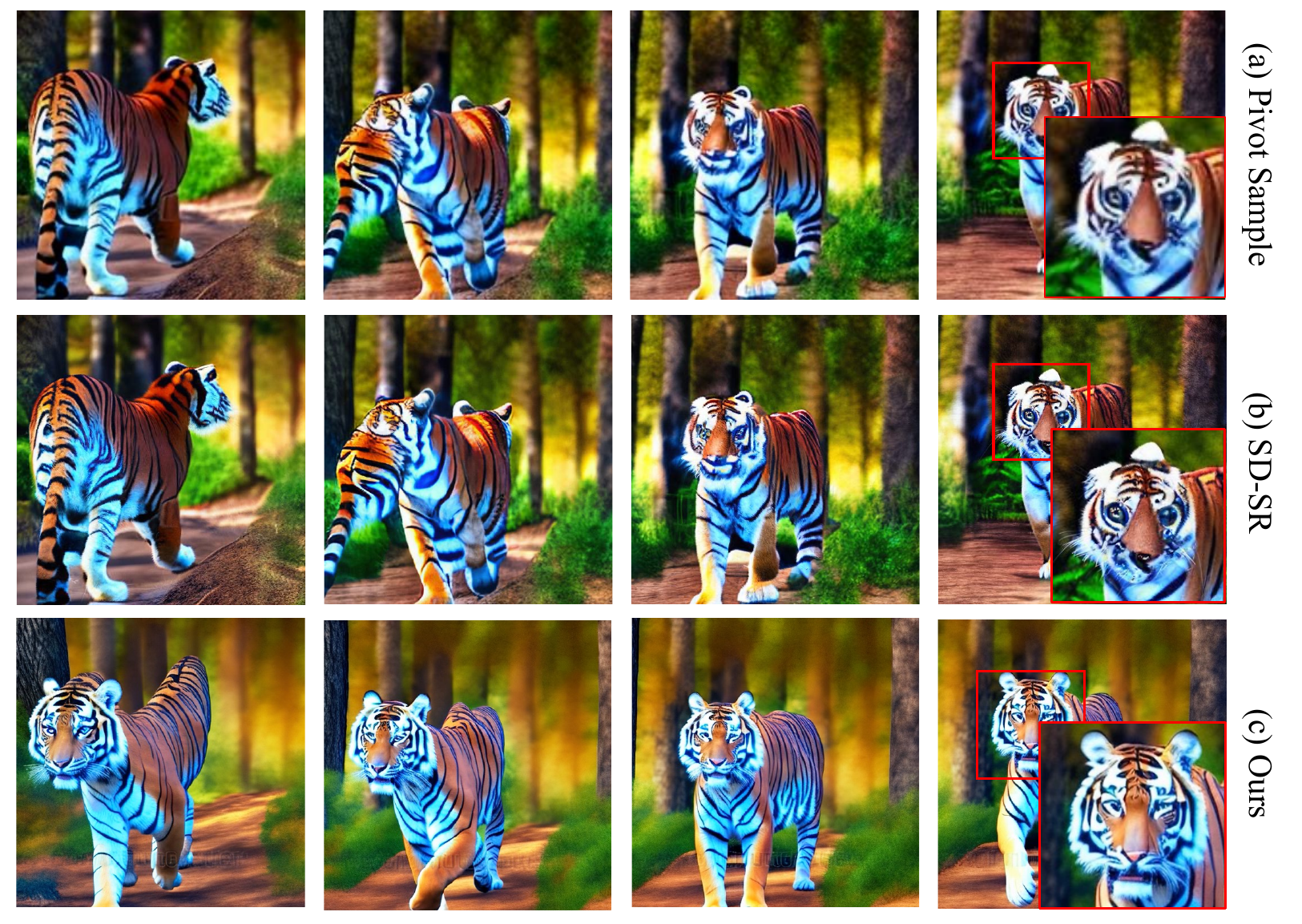}
 \captionof{figure}{Visual examples of video generation of the (a) low-resolution pivot samples generated by the pre-trained base model, (b) super-resolution result by SD+SR, and (c) high-resolution final output of our tuning approach.}
 \vspace{-2mm}
\label{fig:ablation2}
\end{minipage}
\end{table}

\vspace{1mm}
\noindent\textbf{Tuning-free or fine-tuning?}
Although our tuning-free self-cascade diffusion model can inject the correct semantic information to higher-resolution adaptation, some extreme examples still make it difficult to completely suppress repetition issues and composition capabilities, such as repetitive legs and sofas as shown in \cref{fig:ablation1}.
Such failure case is particularly evident in the repetition of very fine-grain objects or texture, which is a common occurrence among all tuning-free competing methods, like Attn-SF~\cite{trainfree-variablesize} and ScaleCrafter~\cite{he2023scalecrafter}.
By tuning plug-and-play and lightweight upsampler modules with a small amount of higher-resolution data, the diffusion model can learn the low-level details at a new scale.

% \noindent\textbf{Number of upsamplers.}
\vspace{1mm}
\noindent\textbf{Relation to the super-resolution methods.}
We also compare our approach to using a pre-trained Stable Diffusion super-resolution (SD 2.1-upscaler-4$\times$) as post-processing, denoted as SD+SR, for the higher-resolution generation as shown in \cref{tab:image_4}.
Our approach achieves better performance and reduced inference time, even in a tuning-free manner (Ours-TF).
In contrast, SD+SR still requires a large amount of high-resolution data for training a new diffusion model with around 184M extra parameters to be trained.
Furthermore, our method not only increases the resolutions of pivot samples like SD+SR, but also \textit{explores the potential of the pre-trained diffusion model for fine-grained details generation and inheriting the composition capacity}. 
We illustrate one example of video generation in \cref{fig:ablation2}, demonstrating two key advantages of our method over SD+SR: (1) while the low-resolution pivot sample from the base model predicts an ``object shift'' result across temporal frames, our method effectively corrects such inconsistencies, which is not achievable by simply applying SD+SR;
(2) our approach excels in synthesizing finer details and textures compared to using SR solely as post-processing, as shown in the enhanced zoom-in on the tiger region.

\vspace{1mm}
\noindent\textbf{Limitations.}
Our method adapts well to higher-resolution domains but still has limitations. 
Since the number of parameters in the upsampler modules we insert is very small, there is an upper bound to the performance of our method when there is sufficient training data, especially when the scale gap is too large, \eg, higher resolution than $4k$ resolution data.
We will further explore the trade-off between adaptation efficiency and generalization ability in future work.

\vspace{-6pt}
\section{Conclusion}
\vspace{-2pt}
In this work, we present a novel self-cascade diffusion model for rapid higher-resolution adaptation. Our approach first introduces a pivot-guided noise re-schedule strategy in a tuning-free manner, cyclically re-utilizing the well-trained low-resolution model. We then propose an efficient tuning version that incorporates a series of plug-and-play, learnable time-aware feature upsampler modules to interpolate knowledge from a small amount of newly acquired high-quality data. Our method achieves over 5x training speed-up with only 0.002M tuning parameters and negligible extra inference time. 
Experimental results demonstrate the effectiveness and efficiency of our approach plugged into various image and video synthesis base models over different scale adaptation settings.

\section*{Acknowledgements}
This research was carried out at the Rapid-Rich Object Search (ROSE) Lab at
Nanyang Technological University in Singapore.
This work was supported in part by the National Research Foundation Singapore Competitive Research Program (award number CRP29-2022-0003), and in part by the National Key R\&D Program of China under grant number 2022ZD0161501. 
\bibliographystyle{splncs04}
\bibliography{main}

\begin{thebibliography}{10}
\providecommand{\url}[1]{\texttt{#1}}
\providecommand{\urlprefix}{URL }
\providecommand{\doi}[1]{https://doi.org/#1}

\bibitem{webvid}
Bain, M., Nagrani, A., Varol, G., Zisserman, A.: Frozen in time: A joint video and image encoder for end-to-end retrieval. In: Proceedings of the IEEE/CVF International Conference on Computer Vision. pp. 1728--1738 (2021)

\bibitem{blattmann2023align}
Blattmann, A., Rombach, R., Ling, H., Dockhorn, T., Kim, S.W., Fidler, S., Kreis, K.: Align your latents: High-resolution video synthesis with latent diffusion models. In: Proceedings of the IEEE/CVF Conference on Computer Vision and Pattern Recognition. pp. 22563--22575 (2023)

\bibitem{bond2023inftydiff}
Bond-Taylor, S., Willcocks, C.G.: $\infty$-diff: Infinite resolution diffusion with subsampled mollified states. arXiv preprint arXiv:2303.18242  (2023)

\bibitem{chai2022any}
Chai, L., Gharbi, M., Shechtman, E., Isola, P., Zhang, R.: Any-resolution training for high-resolution image synthesis. In: European Conference on Computer Vision. pp. 170--188. Springer (2022)

\bibitem{importance-diffusion}
Chen, T.: On the importance of noise scheduling for diffusion models. arXiv preprint arXiv:2301.10972  (2023)

\bibitem{adm}
Dhariwal, P., Nichol, A.: Diffusion models beat gans on image synthesis. Advances in Neural Information Processing Systems  \textbf{34},  8780--8794 (2021)

\bibitem{sd2-1-base}
Diffusion, S.: Stable diffusion 2-1 base. \url{https://huggingface.co/stabilityai/stable-diffusion-2-1-base/blob/main/v2-1_512-ema-pruned.ckpt} (2022)

\bibitem{du2024demofusion}
Du, R., Chang, D., Hospedales, T., Song, Y.Z., Ma, Z.: Demofusion: Democratising high-resolution image generation with no \$\$\$. In: Proceedings of the IEEE/CVF Conference on Computer Vision and Pattern Recognition. pp. 6159--6168 (2024)

\bibitem{gu2023matryoshka}
Gu, J., Zhai, S., Zhang, Y., Susskind, J., Jaitly, N.: Matryoshka diffusion models. arXiv preprint arXiv:2310.15111  (2023)

\bibitem{vqdm}
Gu, S., Chen, D., Bao, J., Wen, F., Zhang, B., Chen, D., Yuan, L., Guo, B.: Vector quantized diffusion model for text-to-image synthesis. In: Proceedings of the IEEE/CVF Conference on Computer Vision and Pattern Recognition. pp. 10696--10706 (2022)

\bibitem{haji2024elasticdiffusion}
Haji-Ali, M., Balakrishnan, G., Ordonez, V.: Elasticdiffusion: Training-free arbitrary size image generation through global-local content separation. In: Proceedings of the IEEE/CVF Conference on Computer Vision and Pattern Recognition. pp. 6603--6612 (2024)

\bibitem{he2023scalecrafter}
He, Y., Yang, S., Chen, H., Cun, X., Xia, M., Zhang, Y., Wang, X., He, R., Chen, Q., Shan, Y.: Scalecrafter: Tuning-free higher-resolution visual generation with diffusion models. arXiv preprint arXiv:2310.07702  (2023)

\bibitem{lvdm}
He, Y., Yang, T., Zhang, Y., Shan, Y., Chen, Q.: Latent video diffusion models for high-fidelity video generation with arbitrary lengths. arXiv preprint arXiv:2211.13221  (2022)

\bibitem{ho2020denoising}
Ho, J., Jain, A., Abbeel, P.: Denoising diffusion probabilistic models. Advances in neural information processing systems  \textbf{33},  6840--6851 (2020)

\bibitem{ho2022cascaded}
Ho, J., Saharia, C., Chan, W., Fleet, D.J., Norouzi, M., Salimans, T.: Cascaded diffusion models for high fidelity image generation. The Journal of Machine Learning Research  \textbf{23}(1),  2249--2281 (2022)

\bibitem{cascaded-dm}
Ho, J., Saharia, C., Chan, W., Fleet, D.J., Norouzi, M., Salimans, T.: Cascaded diffusion models for high fidelity image generation. J. Mach. Learn. Res.  \textbf{23},  47--1 (2022)

\bibitem{simple-diffusion}
Hoogeboom, E., Heek, J., Salimans, T.: simple diffusion: End-to-end diffusion for high resolution images. arXiv preprint arXiv:2301.11093  (2023)

\bibitem{hu2021lora}
Hu, E.J., Shen, Y., Wallis, P., Allen-Zhu, Z., Li, Y., Wang, S., Wang, L., Chen, W.: Lora: Low-rank adaptation of large language models. arXiv preprint arXiv:2106.09685  (2021)

\bibitem{trainfree-variablesize}
Jin, Z., Shen, X., Li, B., Xue, X.: Training-free diffusion model adaptation for variable-sized text-to-image synthesis. arXiv preprint arXiv:2306.08645  (2023)

\bibitem{clean-fid}
Parmar, G., Zhang, R., Zhu, J.Y.: On aliased resizing and surprising subtleties in gan evaluation. In: Proceedings of the IEEE/CVF Conference on Computer Vision and Pattern Recognition. pp. 11410--11420 (2022)

\bibitem{sdxl}
Podell, D., English, Z., Lacey, K., Blattmann, A., Dockhorn, T., M{\"u}ller, J., Penna, J., Rombach, R.: Sdxl: improving latent diffusion models for high-resolution image synthesis. arXiv preprint arXiv:2307.01952  (2023)

\bibitem{radford2021learning}
Radford, A., Kim, J.W., Hallacy, C., Ramesh, A., Goh, G., Agarwal, S., Sastry, G., Askell, A., Mishkin, P., Clark, J., et~al.: Learning transferable visual models from natural language supervision. In: International conference on machine learning. pp. 8748--8763. PMLR (2021)

\bibitem{ldm}
Rombach, R., Blattmann, A., Lorenz, D., Esser, P., Ommer, B.: High-resolution image synthesis with latent diffusion models. In: Proceedings of the IEEE/CVF Conference on Computer Vision and Pattern Recognition. pp. 10684--10695 (2022)

\bibitem{rombach2022high}
Rombach, R., Blattmann, A., Lorenz, D., Esser, P., Ommer, B.: High-resolution image synthesis with latent diffusion models. In: Proceedings of the IEEE/CVF conference on computer vision and pattern recognition. pp. 10684--10695 (2022)

\bibitem{saharia2022palette}
Saharia, C., Chan, W., Chang, H., Lee, C., Ho, J., Salimans, T., Fleet, D., Norouzi, M.: Palette: Image-to-image diffusion models. In: ACM SIGGRAPH 2022 Conference Proceedings. pp. 1--10 (2022)

\bibitem{laion5b}
Schuhmann, C., Beaumont, R., Vencu, R., Gordon, C., Wightman, R., Cherti, M., Coombes, T., Katta, A., Mullis, C., Wortsman, M., Schramowski, P., Kundurthy, S., Crowson, K., Schmidt, L., Kaczmarczyk, R., Jitsev, J.: Laion-5b: An open large-scale dataset for training next generation image-text models (2022)

\bibitem{si2023freeu}
Si, C., Huang, Z., Jiang, Y., Liu, Z.: Freeu: Free lunch in diffusion u-net. arXiv preprint arXiv:2309.11497  (2023)

\bibitem{make-a-video}
Singer, U., Polyak, A., Hayes, T., Yin, X., An, J., Zhang, S., Hu, Q., Yang, H., Ashual, O., Gafni, O., et~al.: Make-a-video: Text-to-video generation without text-video data. arXiv preprint arXiv:2209.14792  (2022)

\bibitem{ddim}
Song, J., Meng, C., Ermon, S.: Denoising diffusion implicit models. arXiv preprint arXiv:2010.02502  (2020)

\bibitem{song2020score}
Song, Y., Sohl-Dickstein, J., Kingma, D.P., Kumar, A., Ermon, S., Poole, B.: Score-based generative modeling through stochastic differential equations. arXiv preprint arXiv:2011.13456  (2020)

\bibitem{su2022dual}
Su, X., Song, J., Meng, C., Ermon, S.: Dual diffusion implicit bridges for image-to-image translation. arXiv preprint arXiv:2203.08382  (2022)

\bibitem{relay-diffusion}
Teng, J., Zheng, W., Ding, M., Hong, W., Wangni, J., Yang, Z., Tang, J.: Relay diffusion: Unifying diffusion process across resolutions for image synthesis. arXiv preprint arXiv:2309.03350  (2023)

\bibitem{FVD}
Unterthiner, T., van Steenkiste, S., Kurach, K., Marinier, R., Michalski, M., Gelly, S.: Towards accurate generative models of video: A new metric \& challenges. arXiv preprint arXiv:1812.01717  (2018)

\bibitem{kvd}
Unterthiner, T., van Steenkiste, S., Kurach, K., Marinier, R., Michalski, M., Gelly, S.: Towards accurate generative models of video: A new metric \& challenges. ICLR  (2019)

\bibitem{wang2023lavie}
Wang, Y., Chen, X., Ma, X., Zhou, S., Huang, Z., Wang, Y., Yang, C., He, Y., Yu, J., Yang, P., et~al.: Lavie: High-quality video generation with cascaded latent diffusion models. arXiv preprint arXiv:2309.15103  (2023)

\bibitem{wu2023tune}
Wu, J.Z., Ge, Y., Wang, X., Lei, S.W., Gu, Y., Shi, Y., Hsu, W., Shan, Y., Qie, X., Shou, M.Z.: Tune-a-video: One-shot tuning of image diffusion models for text-to-video generation. In: Proceedings of the IEEE/CVF International Conference on Computer Vision. pp. 7623--7633 (2023)

\bibitem{difffit}
Xie, E., Yao, L., Shi, H., Liu, Z., Zhou, D., Liu, Z., Li, J., Li, Z.: Difffit: Unlocking transferability of large diffusion models via simple parameter-efficient fine-tuning. arXiv preprint arXiv:2304.06648  (2023)

\bibitem{yu2023video}
Yu, S., Sohn, K., Kim, S., Shin, J.: Video probabilistic diffusion models in projected latent space. In: Proceedings of the IEEE/CVF Conference on Computer Vision and Pattern Recognition. pp. 18456--18466 (2023)

\bibitem{zhang2023show}
Zhang, D.J., Wu, J.Z., Liu, J.W., Zhao, R., Ran, L., Gu, Y., Gao, D., Shou, M.Z.: Show-1: Marrying pixel and latent diffusion models for text-to-video generation. arXiv preprint arXiv:2309.15818  (2023)

\bibitem{any-size-diffusion}
Zheng, Q., Guo, Y., Deng, J., Han, J., Li, Y., Xu, S., Xu, H.: Any-size-diffusion: Toward efficient text-driven synthesis for any-size hd images. arXiv preprint arXiv:2308.16582  (2023)

\end{thebibliography}
\end{document}